\documentclass[journal]{IEEEtran}
\usepackage{framed}
\usepackage[ruled]{algorithm2e}
\usepackage{amssymb}
\usepackage{latexsym}
\usepackage{epsfig}
\usepackage{graphicx}
\usepackage{booktabs,multirow, multicol}
\usepackage{cite}
\usepackage{epsfig}
\usepackage{caption}
 
\usepackage{epstopdf}

\usepackage{url}
\usepackage{xcolor}
\definecolor{newcolor}{rgb}{.8,.349,.1}

\ifCLASSINFOpdf

\fi

\begin{document}

\title{Real-time Eye Gaze Direction Classification Using Convolutional Neural Network}

\author{Anjith~George,~\IEEEmembership{Member,~IEEE,}
        and~Aurobinda~Routray,~\IEEEmembership{Member,~IEEE}
\thanks{ This is a preprint of the accepted paper. To appear in IEEE International Conference on Signal Processing and Communication, SPCOM 2016.}}

\maketitle

\begin{abstract}
Estimation eye gaze direction is useful in various human-computer interaction tasks. Knowledge of gaze direction can give valuable information regarding users point of attention. Certain patterns of eye movements known as eye accessing cues are reported to be related to the cognitive processes in the human brain. We propose a real-time framework for the classification of eye gaze direction and estimation of eye accessing cues. In the first stage, the algorithm detects faces using a modified version of the Viola-Jones algorithm. A rough eye region is obtained using geometric relations and facial landmarks. The eye region obtained is used in the subsequent stage to classify the eye gaze direction.  A convolutional neural network is employed in this work for the classification of eye gaze direction. The proposed algorithm was tested on Eye Chimera database and found to outperform state of the art methods. The computational complexity of the algorithm is very less in the testing phase.  The algorithm achieved an average frame rate of 24 fps in the desktop environment.
\end{abstract}
\IEEEpeerreviewmaketitle

\section{Introduction}

Understanding human emotions and cognitive states are essential in developing a natural human-computer interaction system (HCI). Systems which can identify the affective and cognitive states of humans can make the interaction more natural. The knowledge of mental processes can help computer systems to interact intelligently with humans. Recently, many works have been reported \cite{shan2009facial},\cite{bartlett2003real} investigating the use of facial expression in HCI.  Human eyes provide rich information about human cognitive processes and emotions. The movement patterns of eyes contains information about fatigue \cite{di2012towards}, diseases \cite{terao2011initiation}, etc. Pupil dilation has also been used as an indicator to study cognitive processes \cite{goldinger2012pupil}. The nature of eye movements is unique for each. Recently, several works has been proposed to use eye movement pattern as a biometric \cite{george2015score},\cite{holland2013complex}. 

Most of the works related to facial expression are constrained to desktop environments. However, with the new development of wearable devices like Google Glass \cite{starner2013project} and other augmented reality goggles there are more opportunities for using eye analysis techniques for understanding the affective and cognitive states of the users.

\begin{figure}[h]
\begin{center}
\includegraphics[width=1\linewidth]{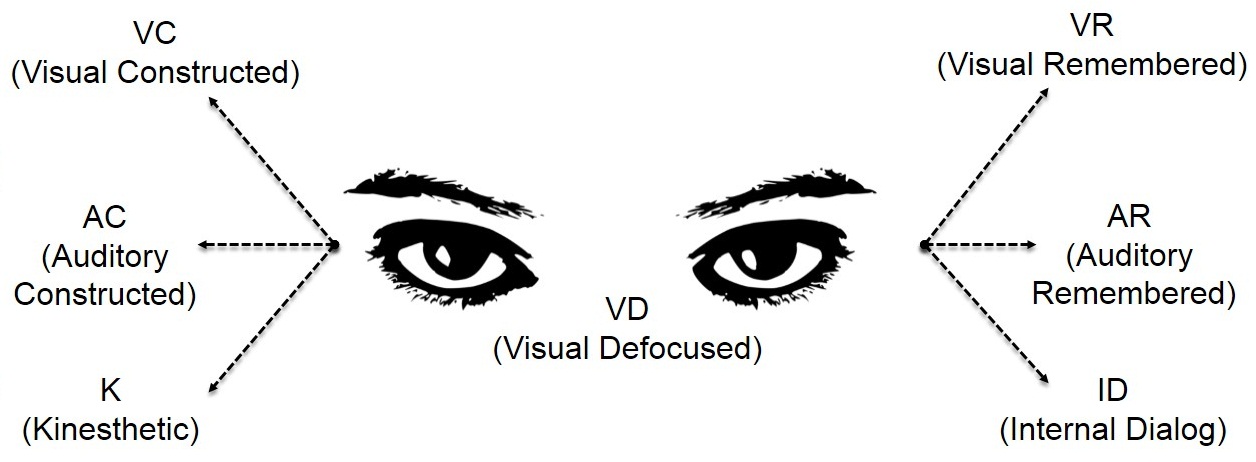}
\end{center}
\caption{Different EACs in NLP theory}
\label{fig:eac}
\end{figure}

The patterns in which the eyes move when humans access their memories is known as eye accessing cues (EAC). The patterns in this non-visual gaze directions have been reported to contain information regarding mental processes. In Neuro-Linguistic Programming (NLP) theory \cite{bandlerfrogs}, eye accessing cues gives information about the mental processes from the direction of eye gaze. These movements are reported to be related to the neural pathways which deal with memory and sensory information. The direction of the iris in the socket can give information regarding various cognitive processes. Each direction of non-visual gaze is associated with different cognitive processes. The meanings of the different EACs are shown in Fig \ref{fig:eac}. More details about EAC model can be found from  \cite{bandlerfrogs}. Even though the EAC theory is not 100 \% accurate, recent studies \cite{sturt2012neurolinguistic},\cite{vranceanu2013computer} have found correlation which encourages further research in the field. A critical review of EAC method can be found in \cite{diamantopoulos2009critical}.

Information retrieval systems can work in a better way if the context is known. Knowledge of the cognitive states can be useful in providing the context in HCI. 

Most of the approaches for gaze estimation uses active infrared based methods which require expensive hardware \cite{duchowski2007eye}. In this paper, we develop a  real-time framework which can detect eye gaze direction using off-the-shelf, low-cost cameras in desktops and other smart devices. Estimation of gaze location from webcam often requires cumbersome calibration procedure \cite{george2016fast}. We treat the gaze direction classification as a multi-class classification problem, avoiding the need for calibration.  The eye directions obtained can be used to find the EAC and thereby infer the user's cognitive process. The information obtained can be useful in the analysis of interrogation videos, human-computer interaction, information retrieval, etc.

The highlights of the paper are shown below:

\begin{itemize}
\item Proposes a real-time framework for eye gaze direction classification
\item We use a Convolutional Neural Network based gaze direction classifier, which is robust against eye localization errors
\item The proposed approach outperforms state of the art algorithms in gaze direction classification
\item The algorithm achieves an average frame rate of 24 fps in desktop environment

\end{itemize}

\section{Related works}

There are many works related to gaze tracking in desktop environments, an excellent review of the methods can be found in \cite{hansen2010eye}. In this section, we limit the discussion
to the recent state of the art works related to eye gaze direction estimation and EAC.

Vr{\^a}nceanu \textit{et al}. proposed a method \cite{vranceanu2011fast} for automatic classification of EAC. The information from color space is used in their approach. The relative position of iris and sclera in the eye bounding box is used to classify the visual accessing cues. Vr{\^a}nceanu \textit{et al}. proposed another method \cite{vranceanu2013automatic} for finding EAC using iris center detection and facial landmark detection. They used isophote curvature based method for iris center localization. The relative position of iris center is used with the fiducial points for a better estimate of eye gaze direction. In \cite{radlak2014gaze} Radlak \textit{et al.} presented a method for gaze direction estimation method in static images. They used an ellipse detector with a Support Vector based verifier. The bounding box is obtained using the hybrid projection functions \cite{song2013literature}. Finally, the gaze direction is classified using Support Vector Machine (SVM) and random forests.


Recently Vr{\^a}nceanu \textit{et al}. \cite{vranceanu2015gaze} proposed another approach for eye direction detection using component separation. Iris, sclera, and skin are segmented and the features obtained are used in a machine learning framework for classifying the eye gaze direction. Zhang \textit{et al}. \cite{zhang2015appearance} applied convolutional neural network for gaze estimation.  They combined the data from face pose estimator and eye region using a CNN model. They have trained a regression model in the output layer.

In most of the related works, the general framework is by using three cascaded stages. Face detection, eye localization, and classification. The localization or classification errors in any of the cascaded stages will result in the reduction of overall accuracy. The computational complexity of the methods is another bottleneck. In this work, we aim at increasing the accuracy of eye gaze direction classification. The developed algorithm is robust against noise, blur, and localization errors. The computational complexity is less in the testing phase, and the proposed algorithm achieves an average 24 fps in a PC based implementation. The proposed framework is described in the following section.

\section{Proposed algorithm}
\begin{figure}[h]
\begin{center}
\includegraphics[width=1\linewidth]{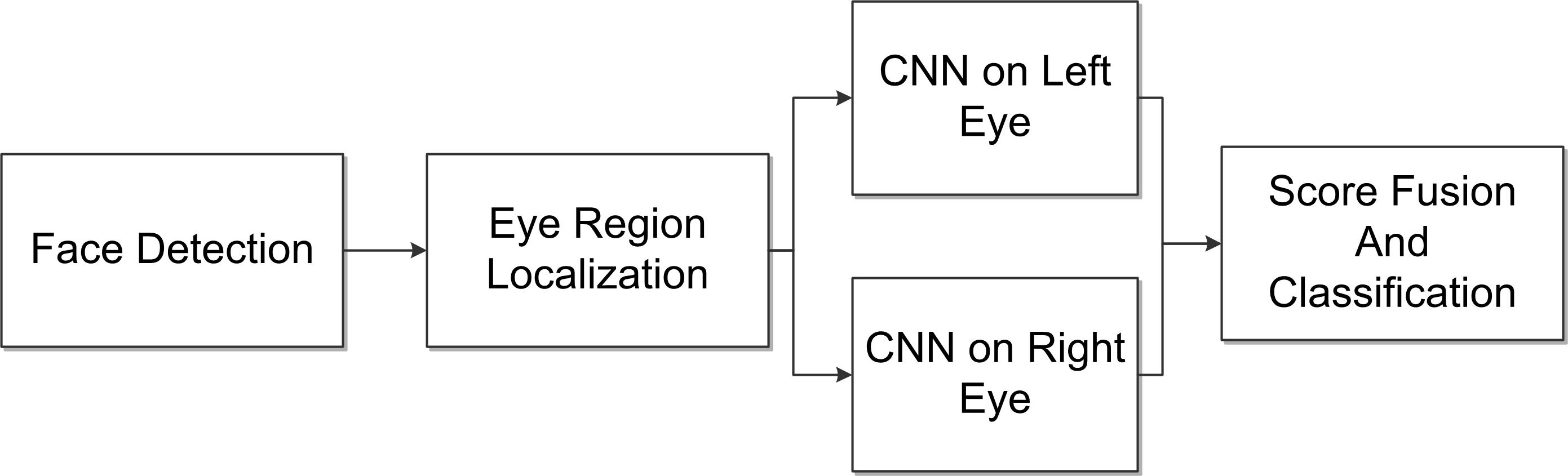}
\end{center}
\caption{Schematic of the overall framework}
\label{fig:overall}
\end{figure}

The overall framework proposed is shown in Fig. \ref{fig:overall}. Different stages of the algorithm are described below.

\subsection{Face detection and eye region localization}
\begin{figure}[h]
\begin{center}
\includegraphics[width=1\linewidth]{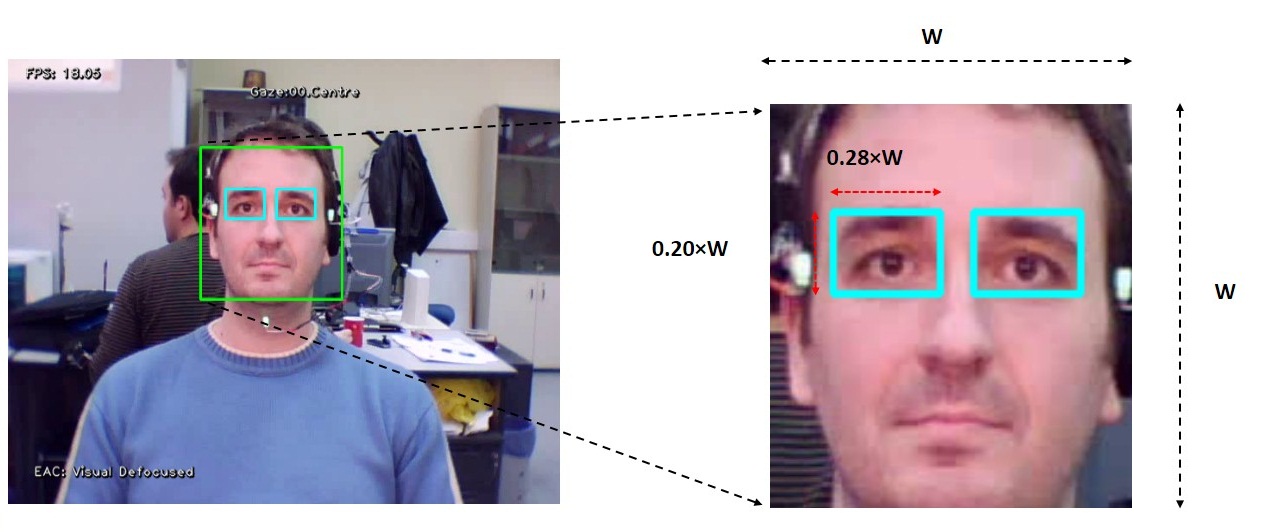}
\end{center}
\caption{Eye region localization using geometric approach (ROI)}
\label{fig:eyeloc}
\end{figure}

The first stage in the algorithm is face detection. We have used a modified version of Viola-Jones 
 method \cite{viola2001rapid} in this paper. The method used is fast and invariant to in-plane rotations, the accuracy and
 trade offs can be found in \cite{dasgupta2013vision}. Once the face region is localized, next stage is to obtain the eye region. We have used two different methods for obtaining the eye region. In the first method, the eye region for classification is obtained geometrically from the face bounding box returned by the face detector (ROI). The dimension of the eye region is shown on an image from HPEG database \cite{asteriadis2009natural} in Fig. \ref{fig:eyeloc}. The eye regions obtained are re-scaled to a resolution of $42\times50$ for the subsequent stages (ROI). In the second method, we used a facial landmark detector to find the eye corners and other fiducial points.

\subsubsection{Facial landmark localization}
Localization of facial landmarks helps in constraining the eye region for classification. Ensemble of Randomized Tree approach (ERT) \cite{kazemi2014one} approach is used for localizing facial landmarks. The face bounding box obtained from the preceding stage is used as the input to the algorithm. The locations of the facial landmarks are regressed using a sparse subset of pixels from the face region. The algorithm is very fast and works even with partial labels. The details of the algorithm can be found in \cite{kazemi2014one}. The eye corner location returned from the landmark detector is used to select the eye region used in the subsequent classification stage.
\subsection{Eye gaze direction classification}

The eye region obtained from the previous stage is used in a multiclass classification framework for
predicting the EAC classes. Convolutional Neural Network (CNN) is used for the classification. The details
of the model used are described below.

\subsubsection{Convolutional Neural Network (CNN)}

\begin{figure*}[!t]
\begin{center}
\includegraphics[width=0.9\linewidth]{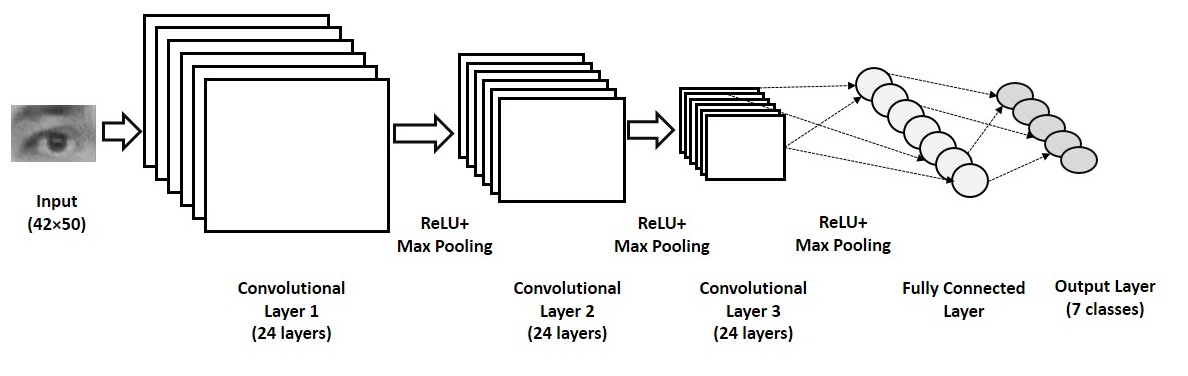}
\end{center}
\caption{Architecture of the CNN  used}
\label{fig:cnn}
\end{figure*}

The convolutional neural network represents a type of feed-forward neural network which can be used for a variety of machine learning tasks. Krizhevsky \textit{at al}. \cite{krizhevsky2012imagenet} used a large CNN model for the classification of images in imagenet database. Even though the training time is huge, the accuracy and robustness of CNNs are better than most of the standard machine learning algorithms. In our approach, we have used a CNN model with three convolution stages. The input stage consists of images of dimension $42\times50$ (or $25\times 15$ in the case of ERT). In the first convolutional layer, 24 filters of dimension $7\times7$ are used. This stage was followed by a rectifier linear unit (ReLU). ReLU layer introduces a non-linearity to the activations. The non-linearity function can be represented as:
\begin{equation}
f(x) = \max (0,x)
\end{equation}
where, $x$ is the input and $f(x)$ the output after the ReLU unit.
A max pooling layer is added after the ReLU stage. Max pooling layer performs a spatial sub-sampling of each output images.
We have used  $2\times 2$ max-pooling layers which reduce the spatial resolution to half. Two similar stages with filter dimensions $5\times5$ and $3\times3$ are also added. After the convolutional, ReLU, and max-pooling layers in the third convolutional layer, the outputs from all the activations are joined in a fully connected layer.
The number of output nodes corresponds to the number of classes in the particular application.  The structure of the network is shown in Fig. \ref{fig:cnn}. The softmax loss is used over classes as the error measure. Cross entropy loss is minimized in the training.

The cross entropy loss ($L$) is defined as:
\begin{equation}
L\left( {f(x),y} \right) =  - y\ln (f(x)) - (1 - y)\ln (1 - fx(x))
\end{equation}

where $x$ is the vector to be classified, $y \in \{ 0,1\} $
, where $y$ is the label

The cross entropy loss is convex and can be minimized using Stochastic Gradient Descent (SGD) \cite{bottou2010large} algorithm. The size of convolution kernels remains same for both ERT and ROI ($7\times7$, $5\times5$ and $3\times3$ ).

\subsubsection{Classification of eye gaze direction}
Two CNN networks are trained independently for left and right eyes. The scores from both the networks are used to obtain the class labels.
\begin{equation}
score = \left( {{{scor{e_L} + scor{e_R}} \over 2}} \right)
\end{equation}
where, $scor{e_L}$ and $scor{e_R}$ denote the scores obtained from left and right CNNs respectively.\\
The class can be found out as the label with maximum probability as:
\begin{equation}
class = \mathop {\arg \max }\limits_{label} (score)
\end{equation}

\section{Experiments}
We have conducted experiments in Eye Chimera database \cite{florea2013can}, \cite{vranceanu2013nlp} which contains all the seven EAC classes. 
This dataset contains images of 40 subjects. For each subject, images with different gaze directions
(for various EAC classes) are available. The total number of images in the dataset is 1170. The ground truth for class labels and fiducial points are also available. 

\subsection{Evaluation procedure}
The database was randomly split into two equal proportions. Training and testing are performed on two completely disjoint 50\% subsets to avoid over-fitting. CNN require a large amount of data in the training phase for better results. The size of the database is relatively small. We have used data augmentation in the training set images to solve this issues. Rotations, blurring and scaling, are performed in the images in the training subset to increase the number of training samples. Two CNNs were
trained separately for left and right eye. In the testing phase, the scores from both left and right eye CNN models are combined to obtain the label of the test image.

We have considered both 7 class and 3 class classification in this work. The methodology followed is same in both the cases.

\subsection{Results}
The results obtained from the experiments in Still Eye Chimera dataset is shown here. The classification 
accuracy was high in 3 class scenario compared to the accuracy in 7 class case.

We have conducted experiments with the two different methods proposed. In the first case, the eye region localization is carried out using geometrical relations. Explicit landmark detection is avoided in this case. This method is denoted as ROI. This approach reduces one stage in the overall framework. Additionally, the robustness of the algorithm against localization errors can be tested. In the second algorithm, we use the ERT based
landmark detection scheme. The eye corners obtained are used to constrain the region for subsequent classification stage. The region obtained in each image is resized to a resolution of $20 \times 15$ for further processing.

In both the cases (ROI and ERT), the data was divided into two 50\% subsets. CNNs were trained separately for left and right eyes using data augmentation. Testing was done 50\% disjoint testing set to avoid over-fitting effects. All the experiments were repeated in both 3 class and 7 class scenario. In 3 class case, we use only classes left, center and right.

The results obtained by using only one eye are shown in Table \ref{tab:componeeye}. 

Combining the information from both eyes improves the accuracy. The results obtained using both the eyes and the comparison with the state of the art is shown in Table \ref{tab:comptwoeye}. 

In both the cases, the proposed method outperforms all the state of the art algorithms in eye gaze direction classification. Highest accuracy is obtained with ERT+CNN algorithm. The individual accuracies achieved in the 7 class case is shown in Table \ref{tab:classacc}.

\begin{figure}[h]
\begin{center}
\includegraphics[width=.8\linewidth]{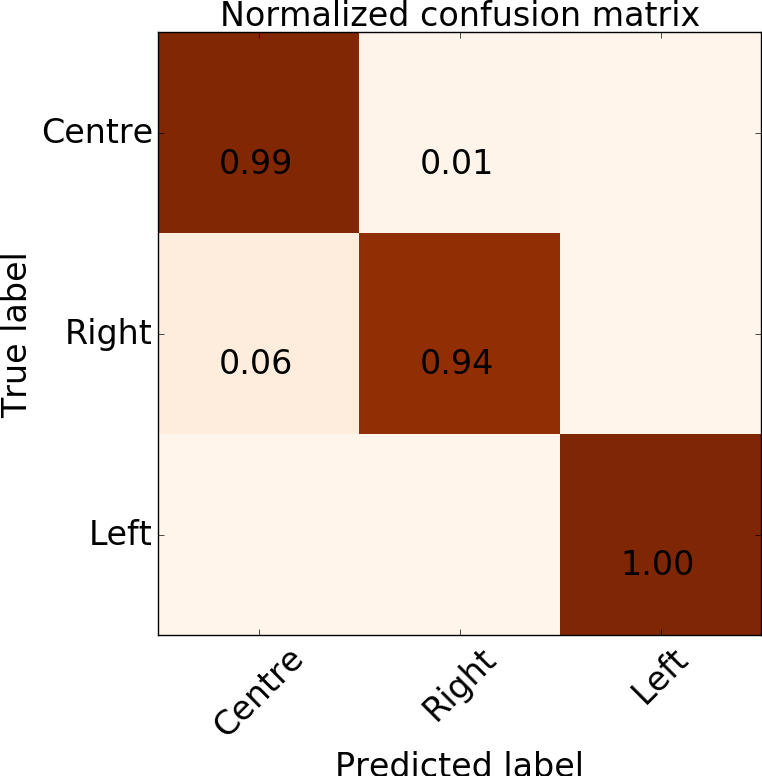}
\end{center}
\caption{Confusion matrix for 3 classes (ERT+CNN)}
\label{fig:conf3}

\end{figure}

\begin{figure}[h]
\begin{center}
\includegraphics[width=1\linewidth]{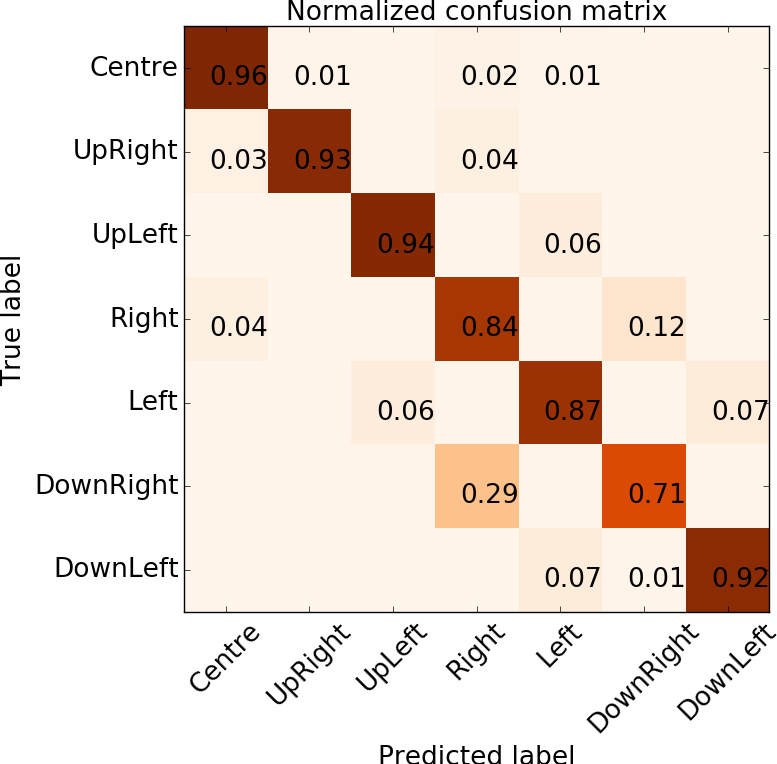}
\end{center}
\caption{Confusion matrix for 7 classes (ERT+CNN)}
\label{fig:conf7}

\end{figure}

The confusion matrix for 3 class and 7 class case (ERT+CNN) are shown in Fig. \ref{fig:conf3} and Fig. \ref{fig:conf7}.
\begin{table}[]
\centering
\caption{Comparison of accuracy of classification using only one eye}
\label{tab:componeeye}
\begin{tabular}{@{}llcc@{}}
\toprule
\begin{tabular}[c]{@{}l@{}}Eye Boundingbox\\ Localization method\end{tabular} & \begin{tabular}[c]{@{}l@{}}Eye direction \\ classification\\ \\ Method\end{tabular} & \begin{tabular}[c]{@{}c@{}}Recognition \\ Rate\\ 7 class (\%)\end{tabular} & \begin{tabular}[c]{@{}c@{}}Recognition \\ Rate\\ 3 class (\%)\end{tabular} \\ \midrule
BoRMaN \cite{valstar2010facial}                                                              & Valenti \cite{valenti2008accurate}                                                                    & 32.00                                                                      & 33.12                                                                      \\
Zhu \cite{zhu2012face}                                                                          & Zhu                                                                                \cite{zhu2012face} & 39.21                                                                      & 45.57                                                                      \\
Vr{\^a}nceanu \cite{vranceanu2015gaze}                                                                      & Vr{\^a}nceanu \cite{vranceanu2015gaze}                                                                                                                                            & 77.54                                                                      & 89.92                                                                      \\
\textbf{\begin{tabular}[c]{@{}l@{}}\textbf{Proposed}\\ (\textbf{Geometric})\end{tabular}}       & \textbf{\begin{tabular}[c]{@{}l@{}}Proposed\\ (CNN)\end{tabular}}                   & \textbf{81.37}                                                             & \textbf{95.98}                                                             \\
\textbf{\begin{tabular}[c]{@{}l@{}}\textbf{Proposed}\\ \textbf{(ERT)}\end{tabular}}             & \textbf{\begin{tabular}[c]{@{}l@{}}Proposed\\ (CNN)\end{tabular}}                   & \textbf{86.81}                                                             & \textbf{96.98}                                                             \\ \bottomrule
\end{tabular}
\end{table}
\begin{table}[]
\centering
\caption{Accuracy in classification  of each class (\%)}
\label{tab:classacc}
\begin{tabular}{@{}llllllll@{}}
\toprule
Method                                                   & \multicolumn{1}{c}{VD} & \multicolumn{1}{c}{VR} & \multicolumn{1}{c}{VC} & \multicolumn{1}{c}{AR} & \multicolumn{1}{c}{AC} & \multicolumn{1}{c}{ID} & \multicolumn{1}{c}{K} \\ \midrule
\begin{tabular}[c]{@{}l@{}}Proposed\\ (ROI)\end{tabular} & 96                     & 79                     & 87                     & 77                     & 79                     & 75                     & 93                    \\
\begin{tabular}[c]{@{}l@{}}Proposed\\ (ERT)\end{tabular} & 97                     & 93                     & 94                     & 84                     & 87                     & 71                     & 91                    \\ \bottomrule
\end{tabular}
\end{table}
\begin{table}[]
\centering
\caption{Accuracy in EAC classification  (\%)  when both the eyes are used}
\label{tab:comptwoeye}
\begin{tabular}{@{}lcccccc@{}}
\toprule
Dataset                                                                      & \multicolumn{1}{l}{Classes} & \multicolumn{1}{l}{\begin{tabular}[c]{@{}l@{}}\cite{valenti2008accurate}\\ +\cite{valstar2010facial}\end{tabular}} & \multicolumn{1}{l}{Zhu \cite{zhu2012face}} & \multicolumn{1}{l}{\cite{vranceanu2015gaze}                                                                     } & \multicolumn{1}{l}{\begin{tabular}[c]{@{}l@{}}\textbf{Proposed}\\ \textbf{(ROI)}\end{tabular}} & \multicolumn{1}{l}{\begin{tabular}[c]{@{}l@{}}\textbf{Proposed}\\ \textbf{(ERT)}\end{tabular}} \\ \midrule
\multirow{2}{*}{\begin{tabular}[c]{@{}l@{}}Still Eye\\ Chimera\end{tabular}} & 7                           & 39.83                                                                            & 43.29                        & 83.08                  & \textbf{85.58 }                                                                       & \textbf{89.81}                                                               \\
                                                                             & 3                           & 55.73                                                                            & 63.01                        & 95.21                  & \textbf{97.65 }                                                                       & \textbf{98.32}                                                               \\ \bottomrule
\end{tabular}
\end{table}
\subsection{Discussion}

The proposed algorithm outperforms all the state of the art results reported in the literature.
From the confusion matrix, it can be seen that most of the mix classifications are in differentiating
between right, down right, etc.  The classification accuracy is poor in the vertical direction (As observed in \cite{vranceanu2015gaze}).
 This can be attributed to the lack of spatial resolution in the vertical 
direction. Most of the cases iris is partly occluded by eyelids in extreme corners. This makes it difficult to classify them accurately. With the larger amount of labeled data, the algorithm could perform even better. Using the color information in the CNN can improve the accuracy even further. A temporal filtering of the predicted labels can improve the accuracy in the case of video.
\section{Conclusion}
In this work,  a framework for real-time classification of eye gaze direction is presented. 
The estimated eye gaze direction is used to infer eye accessing cues, giving information about the cognitive states. The computational complexity is very less; we achieved frame rates around 24 Hz in Python implementation in a 2.0 GHz Core i5 PC running Ubuntu 64 bit (4GB RAM). The per-frame computational time is 42 ms, which is much less than that of the other state of the art methods (250 ms in \cite{vranceanu2015gaze}). Off the shelf webcams can be used for computing the Eye gaze direction.  The proposed algorithm works even with in-plane rotations of the face. The eye gaze direction obtained can also be used for human-computer interaction applications. The computational complexity of the algorithm in testing phase is less, which makes it suitable for smart devices with low-resolution cameras using pre-trained models.
\section{Acknowledgements}
The authors would like to thank Dr. Corneliu Florea for providing the database.
\ifCLASSOPTIONcaptionsoff
  \newpage
\fi
\bibliographystyle{IEEEtran}

\bibliography{refs}

\begin{thebibliography}{10}
\providecommand{\url}[1]{#1}
\csname url@samestyle\endcsname
\providecommand{\newblock}{\relax}
\providecommand{\bibinfo}[2]{#2}
\providecommand{\BIBentrySTDinterwordspacing}{\spaceskip=0pt\relax}
\providecommand{\BIBentryALTinterwordstretchfactor}{4}
\providecommand{\BIBentryALTinterwordspacing}{\spaceskip=\fontdimen2\font plus
\BIBentryALTinterwordstretchfactor\fontdimen3\font minus
  \fontdimen4\font\relax}
\providecommand{\BIBforeignlanguage}[2]{{%
\expandafter\ifx\csname l@#1\endcsname\relax
\typeout{** WARNING: IEEEtran.bst: No hyphenation pattern has been}%
\typeout{** loaded for the language `#1'. Using the pattern for}%
\typeout{** the default language instead.}%
\else
\language=\csname l@#1\endcsname
\fi
#2}}
\providecommand{\BIBdecl}{\relax}
\BIBdecl

\bibitem{shan2009facial}
C.~Shan, S.~Gong, and P.~W. McOwan, ``Facial expression recognition based on
  local binary patterns: A comprehensive study,'' \emph{Image and Vision
  Computing}, vol.~27, no.~6, pp. 803--816, 2009.

\bibitem{bartlett2003real}
M.~S. Bartlett, G.~Littlewort, I.~Fasel, and J.~R. Movellan, ``Real time face
  detection and facial expression recognition: Development and applications to
  human computer interaction.'' in \emph{Computer Vision and Pattern
  Recognition Workshop, 2003. CVPRW'03. Conference on}, vol.~5.\hskip 1em plus
  0.5em minus 0.4em\relax IEEE, 2003, pp. 53--53.

\bibitem{di2012towards}
L.~L. Di~Stasi, R.~Renner, A.~Catena, J.~J. Ca{\~n}as, B.~M. Velichkovsky, and
  S.~Pannasch, ``Towards a driver fatigue test based on the saccadic main
  sequence: A partial validation by subjective report data,''
  \emph{Transportation research part C: emerging technologies}, vol.~21, no.~1,
  pp. 122--133, 2012.

\bibitem{terao2011initiation}
Y.~Terao, H.~Fukuda, A.~Yugeta, O.~Hikosaka, Y.~Nomura, M.~Segawa, R.~Hanajima,
  S.~Tsuji, and Y.~Ugawa, ``Initiation and inhibitory control of saccades with
  the progression of parkinson's disease--changes in three major drives
  converging on the superior colliculus,'' \emph{Neuropsychologia}, vol.~49,
  no.~7, pp. 1794--1806, 2011.

\bibitem{goldinger2012pupil}
S.~D. Goldinger and M.~H. Papesh, ``Pupil dilation reflects the creation and
  retrieval of memories,'' \emph{Current Directions in Psychological Science},
  vol.~21, no.~2, pp. 90--95, 2012.

\bibitem{george2015score}
A.~George and A.~Routray, ``A score level fusion method for eye movement
  biometrics,'' \emph{Pattern Recognition Letters}, 2015.

\bibitem{holland2013complex}
C.~D. Holland and O.~V. Komogortsev, ``Complex eye movement pattern biometrics:
  Analyzing fixations and saccades,'' in \emph{Biometrics (ICB), 2013
  International Conference on}.\hskip 1em plus 0.5em minus 0.4em\relax IEEE,
  2013, pp. 1--8.

\bibitem{starner2013project}
T.~Starner, ``Project glass: An extension of the self,'' \emph{Pervasive
  Computing, IEEE}, vol.~12, no.~2, pp. 14--16, 2013.

\bibitem{bandlerfrogs}
R.~Bandler and J.~Grinder, ``Frogs into princes: Neuro linguistic programming.
  1979.''

\bibitem{sturt2012neurolinguistic}
J.~Sturt, S.~Ali, W.~Robertson, D.~Metcalfe, A.~Grove, C.~Bourne, and
  C.~Bridle, ``Neurolinguistic programming: a systematic review of the effects
  on health outcomes,'' \emph{British Journal of General Practice}, vol.~62,
  no. 604, pp. e757--e764, 2012.

\bibitem{vranceanu2013computer}
R.~Vranceanu, L.~Florea, and C.~Florea, ``A computer vision approach for the
  eye accesing cue model used in neuro-linguistic programming,'' \emph{Sci.
  Bull. Univ. Politehnica Bucharest Ser. C}, vol.~75, no.~4, pp. 79--90, 2013.

\bibitem{diamantopoulos2009critical}
G.~Diamantopoulos, S.~I. Woolley, and M.~Spann, ``A critical review of past
  research into the neuro-linguistic programming eye-accessing cues model,''
  \emph{Current Research in}, p.~8, 2009.

\bibitem{duchowski2007eye}
A.~Duchowski, \emph{Eye tracking methodology: Theory and practice}.\hskip 1em
  plus 0.5em minus 0.4em\relax Springer Science \& Business Media, 2007, vol.
  373.

\bibitem{george2016fast}
A.~George and A.~Routray, ``Fast and accurate algorithm for eye localisation
  for gaze tracking in low resolution images,'' \emph{IET Computer Vision},
  2016.

\bibitem{hansen2010eye}
D.~W. Hansen and Q.~Ji, ``In the eye of the beholder: A survey of models for
  eyes and gaze,'' \emph{Pattern Analysis and Machine Intelligence, IEEE
  Transactions on}, vol.~32, no.~3, pp. 478--500, 2010.

\bibitem{vranceanu2011fast}
R.~Vr{\^a}nceanu, C.~Vertan, R.~Condorovici, L.~Florea, and C.~Florea, ``A fast
  method for detecting eye accessing cues used in neuro-linguistic
  programming,'' in \emph{Intelligent Computer Communication and Processing
  (ICCP), 2011 IEEE International Conference on}.\hskip 1em plus 0.5em minus
  0.4em\relax IEEE, 2011, pp. 225--229.

\bibitem{vranceanu2013automatic}
R.~Vranceanu, C.~Florea, L.~Florea, and C.~Vertan, ``Automatic detection of
  gaze direction for nlp applications,'' in \emph{Signals, Circuits and Systems
  (ISSCS), 2013 International Symposium on}.\hskip 1em plus 0.5em minus
  0.4em\relax IEEE, 2013, pp. 1--4.

\bibitem{radlak2014gaze}
K.~Radlak, M.~Kawulok, B.~Smolka, and N.~Radlak, ``Gaze direction estimation
  from static images,'' in \emph{Multimedia Signal Processing (MMSP), 2014 IEEE
  16th International Workshop on}.\hskip 1em plus 0.5em minus 0.4em\relax IEEE,
  2014, pp. 1--4.

\bibitem{song2013literature}
F.~Song, X.~Tan, S.~Chen, and Z.-H. Zhou, ``A literature survey on robust and
  efficient eye localization in real-life scenarios,'' \emph{Pattern
  Recognition}, vol.~46, no.~12, pp. 3157--3173, 2013.

\bibitem{vranceanu2015gaze}
R.~Vr{\^a}nceanu, C.~Florea, L.~Florea, and C.~Vertan, ``Gaze direction
  estimation by component separation for recognition of eye accessing cues,''
  \emph{Machine Vision and Applications}, vol.~26, no. 2-3, pp. 267--278, 2015.

\bibitem{zhang2015appearance}
X.~Zhang, Y.~Sugano, M.~Fritz, and A.~Bulling, ``Appearance-based gaze
  estimation in the wild,'' in \emph{2015 IEEE Conference on Computer Vision
  and Pattern Recognition}.\hskip 1em plus 0.5em minus 0.4em\relax IEEE
  Computer Society, 2015.

\bibitem{viola2001rapid}
P.~Viola and M.~Jones, ``Rapid object detection using a boosted cascade of
  simple features,'' in \emph{Computer Vision and Pattern Recognition, 2001.
  CVPR 2001. Proceedings of the 2001 IEEE Computer Society Conference on},
  vol.~1.\hskip 1em plus 0.5em minus 0.4em\relax IEEE, 2001, pp. I--511.

\bibitem{dasgupta2013vision}
A.~Dasgupta, A.~George, S.~Happy, and A.~Routray, ``A vision-based system for
  monitoring the loss of attention in automotive drivers,'' \emph{Intelligent
  Transportation Systems, IEEE Transactions on}, vol.~14, no.~4, pp.
  1825--1838, 2013.

\bibitem{asteriadis2009natural}
S.~Asteriadis, D.~Soufleros, K.~Karpouzis, and S.~Kollias, ``A natural head
  pose and eye gaze dataset,'' in \emph{Proceedings of the International
  Workshop on Affective-Aware Virtual Agents and Social Robots}.\hskip 1em plus
  0.5em minus 0.4em\relax ACM, 2009, p.~1.

\bibitem{kazemi2014one}
V.~Kazemi and J.~Sullivan, ``One millisecond face alignment with an ensemble of
  regression trees,'' in \emph{Computer Vision and Pattern Recognition (CVPR),
  2014 IEEE Conference on}.\hskip 1em plus 0.5em minus 0.4em\relax IEEE, 2014,
  pp. 1867--1874.

\bibitem{krizhevsky2012imagenet}
A.~Krizhevsky, I.~Sutskever, and G.~E. Hinton, ``Imagenet classification with
  deep convolutional neural networks,'' in \emph{Advances in neural information
  processing systems}, 2012, pp. 1097--1105.

\bibitem{bottou2010large}
L.~Bottou, ``Large-scale machine learning with stochastic gradient descent,''
  in \emph{Proceedings of COMPSTAT'2010}.\hskip 1em plus 0.5em minus
  0.4em\relax Springer, 2010, pp. 177--186.

\bibitem{florea2013can}
L.~Florea, C.~Florea, R.~Vr{\^a}nceanu, and C.~Vertan, ``Can your eyes tell me
  how you think? a gaze directed estimation of the mental activity,'' in
  \emph{Proceedings of the British Machine Vision Conference. BMVA Press},
  2013, pp. 60--1.

\bibitem{vranceanu2013nlp}
R.~Vr{\^a}nceanu, C.~Florea, L.~Florea, and C.~Vertan, ``Nlp eac recognition by
  component separation in the eye region,'' in \emph{Computer Analysis of
  Images and Patterns}.\hskip 1em plus 0.5em minus 0.4em\relax Springer, 2013,
  pp. 225--232.

\bibitem{valstar2010facial}
M.~Valstar, B.~Martinez, X.~Binefa, and M.~Pantic, ``Facial point detection
  using boosted regression and graph models,'' in \emph{Computer Vision and
  Pattern Recognition (CVPR), 2010 IEEE Conference on}.\hskip 1em plus 0.5em
  minus 0.4em\relax IEEE, 2010, pp. 2729--2736.

\bibitem{valenti2008accurate}
R.~Valenti and T.~Gevers, ``Accurate eye center location and tracking using
  isophote curvature,'' in \emph{Computer Vision and Pattern Recognition, 2008.
  CVPR 2008. IEEE Conference on}.\hskip 1em plus 0.5em minus 0.4em\relax IEEE,
  2008, pp. 1--8.

\bibitem{zhu2012face}
X.~Zhu and D.~Ramanan, ``Face detection, pose estimation, and landmark
  localization in the wild,'' in \emph{Computer Vision and Pattern Recognition
  (CVPR), 2012 IEEE Conference on}.\hskip 1em plus 0.5em minus 0.4em\relax
  IEEE, 2012, pp. 2879--2886.

\end{thebibliography}

\end{document}